\documentclass[10pt,twocolumn,letterpaper]{article}

\usepackage{cvpr}
\usepackage{times}
\usepackage{epsfig}
\usepackage{graphicx}
\usepackage{amsmath}
\usepackage{amssymb}
\usepackage{multirow}
\usepackage{float}
\usepackage{color}
\usepackage{array}
\usepackage{mathrsfs}
\usepackage{tabularx}
\usepackage{bm}

\usepackage[breaklinks=true,bookmarks=false]{hyperref}

\cvprfinalcopy 

\begin{document}
	
	\title{Complex Sequential Understanding through the Awareness of Spatial and Temporal Concepts}
	\author{Bo Pang, Kaiwen Zha, Hanwen Cao, Jiajun Tang, Minghui Yu, Cewu Lu*\\
		Shanghai Jiao Tong University\\
		{\tt\small \{pangbo, Kevin\_zha, mbd\_chw, yelantingfeng, 1475265722, lucewu\}@sjtu.edu.cn}}

	\maketitle
	\renewcommand{\thefootnote}{\fnsymbol{footnote}}
	
	\begin{abstract}
		Understanding sequential information is a fundamental
		task for artificial intelligence. Current neural
		networks attempt to learn spatial and temporal information as a whole, limited their abilities to represent large scale spatial representations over long-range sequences. Here, we introduce a new modeling strategy called Semi-Coupled Structure (SCS), which	consists of deep neural networks that decouple the complex
		spatial and temporal concepts learning. Semi-Coupled Structure can learn to implicitly separate input information into independent parts and process
		these parts respectively. Experiments demonstrate that a Semi-Coupled Structure can successfully annotate the outline of an object in images sequentially and perform video action recognition. For sequence-to-sequence problems, a Semi-Coupled Structure can predict future meteorological radar echo images based on observed images. Taken together, our results demonstrate that a Semi-Coupled Structure has the capacity to improve the performance of LSTM-like models on large scale sequential tasks.
	\end{abstract}
	
	Complex sequential tasks involve extremely high-dimensional spatial signal over long timescales. Neural networks have made breakthroughs in sequential learning~\cite{graves2013generating,sutskever2014sequence}, visual understanding~\cite{krizhevsky2012imagenet,he2016deep,he2017mask}, and robotic tasks~\cite{levine2016end,schulman2015trust}. 
	Conventional neural networks treat spatial and temporal information as a whole, processing these parts together. This limits their ability to solve complex sequential tasks involving high-dimensional spatial and temporal components~\cite{feichtenhofer2018slowfast,kim2017residual}. A natural idea to address this limitation is to learn the two different concepts relatively independently.
	
	Here, we introduce a structure that decouples spatial and temporal information, implicitly learning respective spatial and temporal concepts through a deep comprehensive model. We find that such concept decomposition significantly simplifies the learning and understanding process of complex sequences.
	Due to the differentiable property of this structure, which we call Semi Coupled Structure (SCS), we can train it end to end with gradient descent, allowing it to effectively learn to decouple and integrate information in a goal-directed manner.
	
	\begin{figure*}
		\begin{center}
			\includegraphics[width=\linewidth]{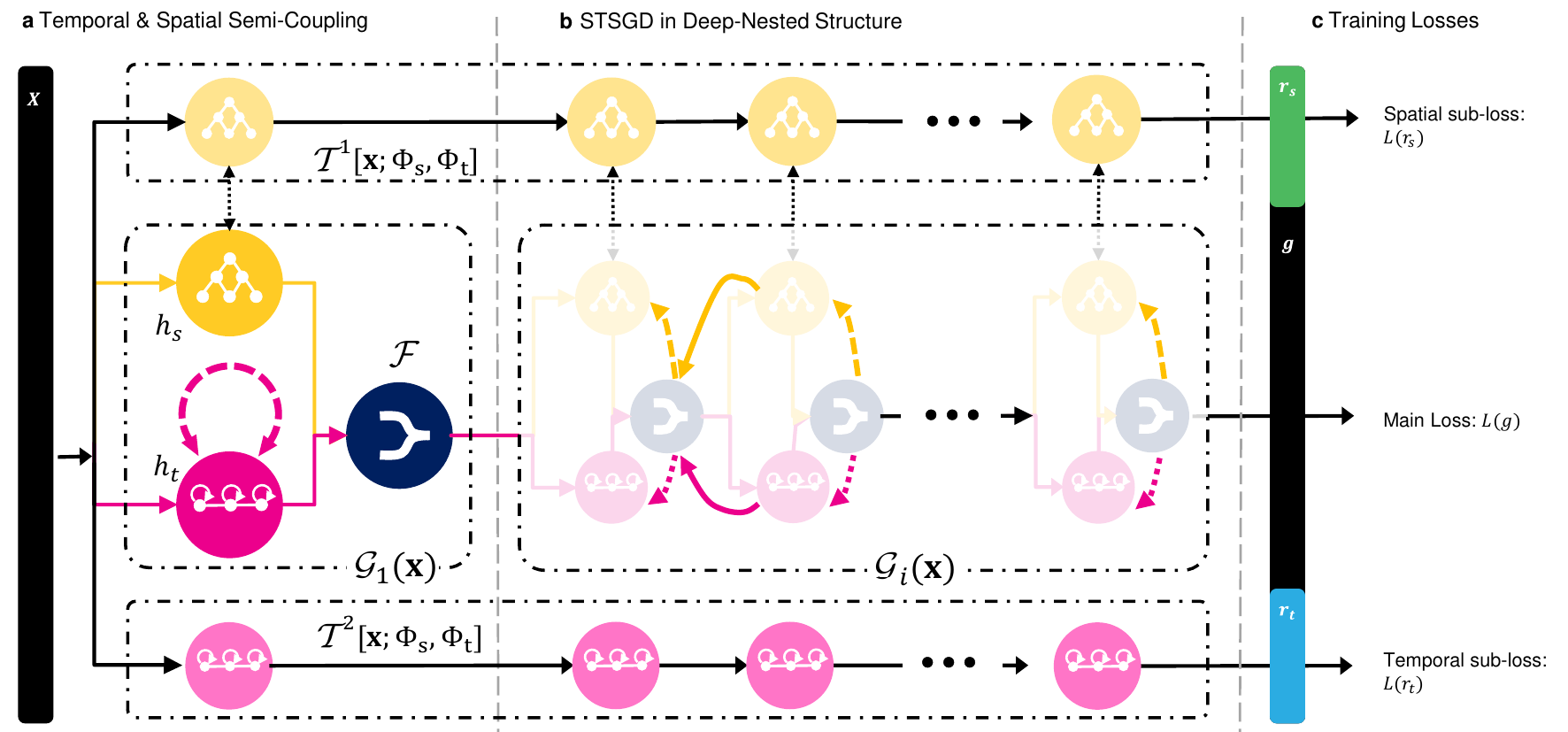}
		\end{center}
		\caption{\textbf{The whole pipeline of our Semi-Coupled Structure}. \textbf{a}, With input $\mathbf{x}$, $\mathcal{G}$ decouples the spatial-temporal information by $h_t$ which focuses on temporal features, $h_s$ that mainly extracts spatial features, and $\mathcal{F}$ integrates them to form the complete temporal-spatial semi-coupled system. \textbf{b}, To keep the semi-coupled peculiarity in a deep structure, we design the Spatial-Temporal Switch Gradient Descent (STSGD) method (see Sec.~\ref{sec:STSGD}) that stops the gradient back propagating through the dashed lines in a certain probability $p$ to decouple the training processes of $h_s$ and $h_t$. $\mathcal{T}^1$ and $\mathcal{T}^2$ are utilized to make $h_s({\cdot})$ and $h_t(\cdot)$ further focus on their own roles and monitor the training schedule of $h_s$ to adjust $q$ in Advanced STSGD (ASTSGD). \textbf{c}, Except the main training loss ($L(g)$) of Semi-Coupled Structure based on main goal $g$, there are another two losses $L(r_s)$, $L(r_t)$ based on sub-goals $r_s$, $r_t$ for $\mathcal{T}^1$, and $\mathcal{T}^2$ to guide $h_s$ and $h_t$ to focus on spatial and temporal features respectively.}
		\label{fig:pipeline}
	\end{figure*}
	
	\section{Awareness of Spatial and Temporal Concept}\label{sec:aware}
	In the brain, there are two different pathways that feed temporal information and contextual representations respectively into the hippocampus~\cite{kitamura2015entorhinal}. This implies that spatial and temporal concepts are learnt by different cognitive mechanisms and, moreover, that they should be synchronized in order to effectively process sequential information. Taking inspiration from this mechanism in the brain, the deep neural model that is implicitly aware of the two concepts can be formulated as:
	\begin{align}
	\mathcal{F}[h_s(\mathbf{x}|\psi_s), h_t(\mathbf{x}|\psi_t)] \label{eq:split}
	\end{align}
	where $\mathbf{x}$ is the input, and $\psi_s$ and $\psi_t$ are the parameters to optimize. $h_s$ aims at extracting spatial information, while $h_t$ is designed to handle temporal learning. These two kinds of information are fed into $\mathcal{F}$ which is designed to output the final processing results, just like the hippocampus. 
	
	\begin{figure*}
		\begin{center}
			\includegraphics[width=\linewidth]{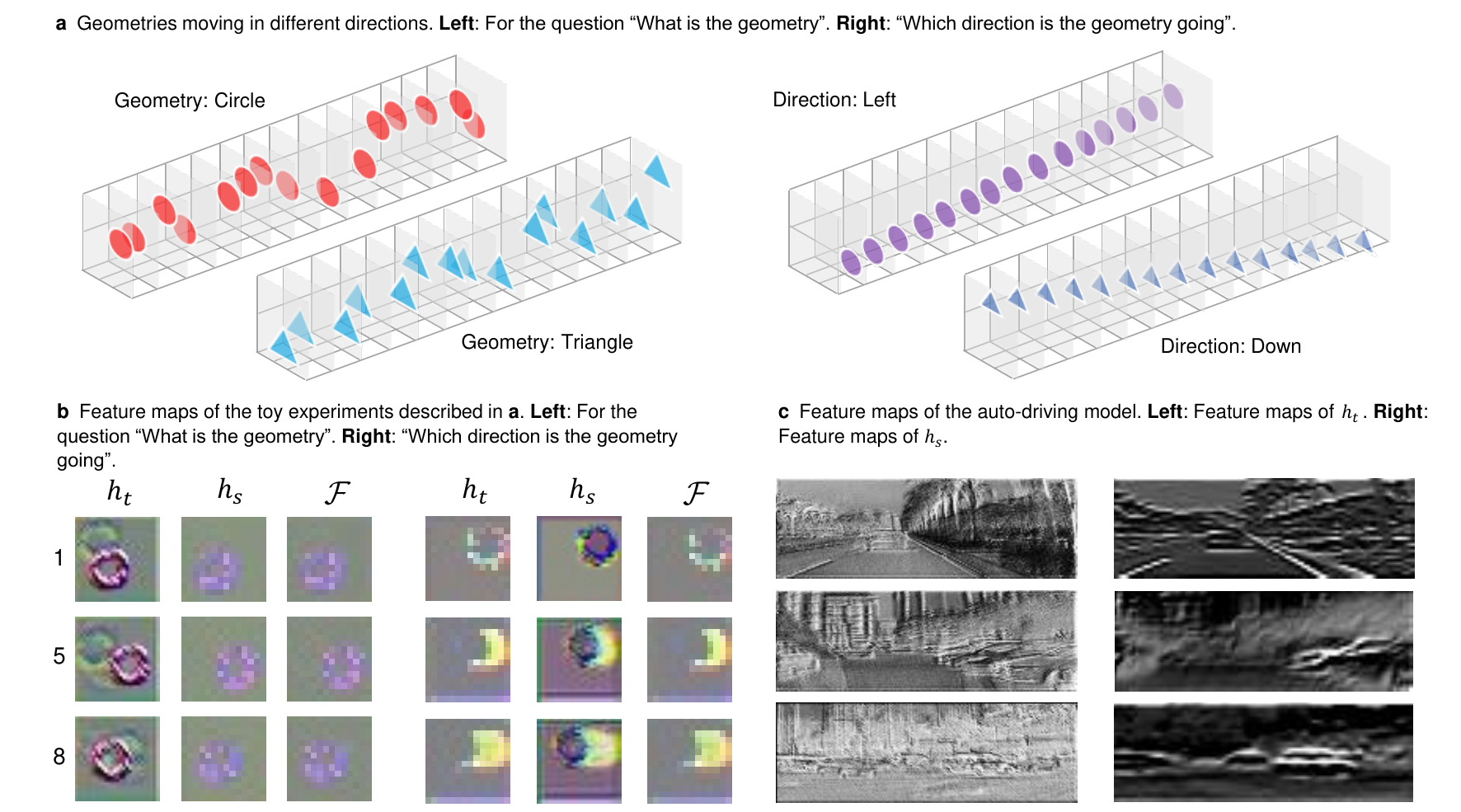}
		\end{center}
		\caption{\textbf{Toy experiments for semi-coupling Structure}. \textbf{a}, The toy examples designed to demonstrate the SCS scheme can successfully decouple the temporal-spatial features. The contents of input sequences are moving geometries. We let the model to distinguish the shapes (left two sequences) and the moving directions (right two sequences). \textbf{b}, The feature maps from $h_s$, $h_t$ and $\mathcal{F}$ in the top layer of the model on the tasks described in \textbf{a}. When distinguish the geometry's shape (left three columns), we can see that $h_s$ and $\mathcal{F}$ only contain spatial information, and temporal features in $h_t$ do not intrude into $h_s$ in high layers due to the filter function of $\mathcal{F}$ in low layers. While for the task to distinguish the moving directions, we can see that the temporal information is integrated into $\mathcal{F}$ and the spatial features in $h_s$ are weakened. \textbf{c}, Feature maps of $h_s$ and $h_t$ on auto driving tasks. The highlighting parts in the feature maps of $h_t$ describe more information about the scene changing, while the highlighting parts in $h_s$ focus on the outline of the objects and roads.}
		\label{fig:toyexample} 
	\end{figure*}

	We further advance our model by considering the fact that spatial and temporal information are deeply coupled with each other, when processed by a brain~\cite{oliveri2009spatial}. Therefore, the model can naturally be extended as a deep nested structure to model such mutual-coupling. We define the $i^{th}$ coupling unit as:
	\begin{align}
	\mathcal{G}_i(\mathbf{x}_i) = \mathcal{F}[h_s(\mathbf{x}_i|\psi_s^i), h_t(\mathbf{x}_i|\psi_t^i)] \label{eq:G}
	\end{align}
	thus, the deep spatial-temporal Semi-Coupled Structure can be expressed as:
	\begin{align}
	\mathcal{T}[\mathbf{x};\Psi_s, \Psi_t] = \mathcal{G}_1\circ \mathcal{G}_2 \circ ... \circ \mathcal{G}_n(\mathbf{x}) \label{eq: Gs}
	\end{align}
	where $n$ is the depth of the deep nested model, and $\Psi_s = \{\psi_s^1,...,\psi_s^n\}$ and $\Psi_t =\{\psi_t^1,...,\psi_t^n\}$ are the parameter sets. 
	
	In this structure, spatial and temporal information are intertwined deeply and collaboratively, meanwhile, $h_s(\cdot)$ and $h_t(\cdot)$ are responsible for spatial and temporal concept processing respectively. To this end, we propose two design paradigms (see Fig.~\ref{fig:pipeline}).

	\begin{itemize}
		\item \textbf{Structure Paradigms}~ $h_s(\cdot)$ and $h_t(\cdot)$ work as their roles by their different structural designs. At a certain time stamp of the sequence, the structure of $h_t(\cdot)$ should have access to the temporal information of other stamps in the sequence like the Recurrent Neural Network (RNN). While for $h_s(\cdot)$, it has no direct connection to the samples of other time stamps, so it can focus on the spatial information, which normally can be a CNN structure. 
		
		\item \textbf{Task Paradigms}~ Because of the deep nested structure, $h_s(\cdot)$ and $h_t(\cdot)$ will disturb each other. To make $h_s(\cdot)$ and $h_t(\cdot)$ further focus on their roles, besides the main goal: $g=\mathcal{T}[\mathbf{x}; \Psi_s, \Psi_t]$, we assign two extra sub-goals: $r_s = \mathcal{T}^1[\mathbf{x};\Psi_s, \Psi_t]$ and $r_t = \mathcal{T}^2[\mathbf{x};\Psi_s, \Psi_t]$, where $\mathcal{T}^1$ and $\mathcal{T}^2$ share the same model components and parameters with $\mathcal{T}$. We also call $r_s$ and $r_t$ as the spatial and temporal indicating goals which can reflect the qualities of spatial and temporal features. The key design is to make both two indicating goals only impact on their own parameters: $\Psi_s$ or $\Psi_t$. That is, in $\mathcal{T}^1$, ${\partial r_s} / {\partial \Psi_t} = 0$ and in $\mathcal{T}^2$, ${\partial r_t} / {\partial \Psi_s} = 0$. The specific definition of the sub-goals depends on different tasks. Taking action recognition as an example, $r_s$ can be human poses in a single frame that is unrelated to temporal information but useful for action understanding, and $r_t$ can be the estimates of optical flow.
	\end{itemize}
	This new modeling strategy is called Semi-Coupled Structure (SCS). It is a general framework that is easy to be revised to fit various applications. If the temporal indicating label of a specific application is difficult to provide, we find that only $r_s$ is enough to encourage $h_s(\cdot)$ to focus on spatial learning, and $h_t(\cdot)$ can naturally take the responsibility of the remain (temporal) information. 
	
	\paragraph{Discussion} The proposed SCS makes each component be responsible for a specific sub-concept (spatial or temporal). This strategy widely exists in the brain using several different encephalic regions to complete a single complex task~\cite{wolman2012tale,diez2015novel}. During this process, our method learns to separate temporal and spatial information, even though we do not define them separately. There are previous works that also try to separate the temporal and spatial information, but they adopt the hand-craft spatial and temporal definitions, like Two-Stream model~\cite{simonyan2014two} which uses optical flow~\cite{lucas1986generalized} to define temporal information and SlowFast Networks~\cite{feichtenhofer2018slowfast} that uses asymmetrical spatial and temporal sampling density to distinguish and define them. Fig.~\ref{fig:toyexample} illustrates that SCS can successfully decouple the temporal and spatial information in visual sequences. In a toy experiment, the visual sequences show different geometries moving in different directions (see Fig.~\ref{fig:toyexample} \textbf{a} for details) and we note that our method outperforms LSTM~\cite{hochreiter1997long} on recognizing ``Which direction is the geometry going?" when it encounters a specific geometry it never saw and ``what is the geometry?" when the motion is disordered. Fig.~\ref{fig:toyexample} \textbf{b} shows the feature maps of $h_t(\mathbf{x})$ and $h_s(\mathbf{x})$, and we can primarily recognize that $h_t(\mathbf{x})$ represents temporal-related features and $h_s(\mathbf{x})$ is for the spatial one from the viewpoint of human vision. Interestingly, our model can quantitatively indicate how important the temporal information is toward the final goal by comparing the indicating goals $r_s$ and $r_t$, thanks to the awareness of the temporal and spatial concepts. We believe this quantitative indicator will largely benefit the sequential analysis.
	
	\section{Performance Profiling on Academic and Reality Datasets}
	\subsection{Video action recognition experiments}
	To investigate the capacity of the Semi-Coupled Structure, we conduct the first experiment on video action recognition task. We choose UCF-101~\cite{soomro2012ucf101}, HMDB-51~\cite{kuehne2011hmdb} and Kinetics-400~\cite{carreira2017quo} datasets which consist of short videos describing human actions collected from website. Correct classification is inseparable from the comprehensive abilities of extracting temporal and spatial information: for example, distinguishing ``triple jump" and ``long jump" requires a structure to precisely understand temporal information, while to tell ``sweeping floor" and ``mopping floor" apart requires great spatial information processing ability. We find that our SCS can successfully learn the temporal information over long-range sequences on limited resources and compared to the conventional sequential models, such as LSTM stacked on CNN~\cite{donahue2015long} and ConvLSTM~\cite{xingjian2015convolutional}, our SCS achieves remarkable improvements (see Tab.~\ref{tab:actionResult} for details). Unlike the previous architectures, our SCS can be trained end-to-end without the support of a backbone network (such as VGG~\cite{simonyan2014very}, ResNet~\cite{he2016deep}, and Inception~\cite{szegedy2015going}. Due to the temporal and spatial semi-coupling, the network can reduce the interference from the temporal unit to the spatial one so that we can still get high-quality spatial features.
	
	\subsection{Object outline sequentially annotation experiments}
	Although the video action recognition task takes a sequence as input, each sequence only need to be assigned one action label. Therefore, modeling it as a pattern recognition problem instead of a sequence learning is also a way to go. For example, 3D convolution model~\cite{ji20133d,carreira2017quo} is widely used recently. Based on this consideration, we need a typical sequential task to further validate the SCS's performance. We, therefore, turn to the outline annotation task.
	
	Unlike video action recognition, outline annotation task calls for a point sequence to represent the outline of the target. Each input consists of an image with a start point to declare which object is the annotation target and an end point to indicate which direction to annotate. The annotation models are trained to give out the outline's key points of the target object one by one from the provided start point to the end point. A new key point is generated based on the already calculated key points (Fig.~\ref{fig:experiments} a). The generated key points form the predicted outline and we adopt the IoU between the predicted and ground-truth outline as the evaluation metric.
	Because it is not easy to give out the complete sequential key points in one step only with the start and end point, it is not suitable to model this task as a pattern recognition task like the video action recognition task.
	
	We adopt CityScapes dataset~\cite{cordts2016cityscapes} as our data source and the target objects are all from the outdoor scene. The relatively complex backgrounds require great ability to extract spatial features. Different from the action recognition task, the temporal information lies in the sequential key point positions which act as the attentions to assist the selection of the subsequent points. As a benchmark we compare our SCS based model, a modified Polygon-RNN model~\cite{castrejon2017annotating}, with the original LSTM based Polygon-RNN model. In this case, our deep SCS model reaches an average of 70.4 in terms of IoU, 15\% relative improvements over the baseline. Fig.~\ref{fig:experiments} c illustrates the training processes of SCSs with different depths and training strategies.
	
	\subsection{Auto-driving experiments}
	Next, we want to evaluate the performance of SCS on some cutting-edge applications. Still, we start from a pattern recognition like problem: the simplified auto-driving problem. We treat the problem as a visual sequence processing task so that we only focus on the driving direction and ignore the route planning, strong driving safety and other things in the real driving environments.
	
	A driving agent, given the sequence of driver's perspective images, needs to decide the driving direction for the last image. It is worth noting that the agent does not know the historical direction to avoid it making ``lazy decision": simply repeating the recent direction. As the previous experiments, this task also requires great ability to process spatial information to figure out the road direction and obstruction condition, and ability to capture temporal information to make coherent decisions.
	
	We evaluate the SCS on the Comma.ai dataset~\cite{santana2016learning} and the LiVi dataset~\cite{chen2018lidar}. The image sequences are the driving videos in real traffic including varied scenes such as highways and mountain roads, and the behaviours of the driver are recorded as the direction label. By experiments, we find that features from $h_s$ record more features of the current road and $h_t$ records more about scene changes during driving (Fig.~\ref{fig:toyexample} c). This indicates that they divide the works successfully and just as the design purpose, they focus on temporal and spatial features respectively so SCS can remarkably disperse the learning pressure to different components. Again, the SCS performs substantially better than conventional LSTM models (see Tab.~\ref{tab:drivingResult}).
	
	\subsection{Precipitation forecasting experiments}
	We further apply our SCS model to precipitation forecasting task in order to test its performance on sequence generation problem. Unlike the previous experiments, where the model receives the input sequence and gives out the output sequence synchronously, we apply a form of ``sequence to sequence" learning~\cite{sutskever2014sequence} in which the input sequence is encoded into a representation and then the model gives out the output sequence based on this representation.
	
	Our dataset, which we term as REEC-2018, contains a set of meteorological Radar Echo images for Eastern China in 2018. The metric of the radar echo is composite reflectivity (CR)  which can be utilized to predict the precipitation intensity. A model, given a sequence of the radar echo images sorted in time, needs to predict a sequence of the future radar echo images from the previous evolution of CR (see Fig.~\ref{fig:experiments} b).
	
	Through experiments, we find that our SCS model can successfully generate the results with original evolution trends, such as diffusion and translation. Compared with the ConvLSTM, a conventional sequence model for visual, again, our SCS gains huge performance improvements.
	
	\section{Discussion}
	In summary, we have built a Semi-Coupled Structure that can learn to divide the work of extracting features automatically. A major reason for utilizing such a structure is to alleviate the interference between learning temporal and spatial features. Many techniques like STSGD and LTSC (see Methods and Fig.~\ref{fig:STSGD}) are proposed to make the SCS easier to train. The performances of our structure are provided by the experiments, and the theme connecting these experiments is the need to synthesize high dimensional temporal and spatial features embedded in data sequences. All the experiments demonstrate that SCS is able to process visual sequential data regardless of whether the task is sensory processing or sequence learning. Moreover, we have seen that the temporal and spatial features are handled separately by different sub-structures due to the temporal-spatial semi-coupling mechanism (see Fig.~\ref{fig:toyexample} and Fig.~\ref{fig:ablation_study}).
	
	\section{Related Work}
		\paragraph{Sequence models}
		Sequential tasks on high dimensional signal require a model to extract spatial representations as well as temporal features. A series of prior works has shed light on these tough problems: Constrained by the computational resource, inchoate methods~\cite{karpathy2014large,yue2015beyond,wang2016actionness,weinzaepfel2015learning} do not explicitly extract temporal feature, instead, acquire global features by combining spatial information, where pooling is a common method. To extract temporal information, some researchers adopt low-level features, like optical flow~\cite{simonyan2014two,carreira2017quo}, trajectories~\cite{wang2011action,wang2013dense}, and pose estimation~\cite{maji2011action} to deal with temporal information. These low-level features are easy to extract but they are handcraft to some extent, therefore, the performance is limited. Then with more computational resource, Recurrent Neural Networks (RNN)~\cite{donahue2015long,wu2015modeling,srivastava2015unsupervised} are widely used, where hidden states take charge of ``remembering" the history and extract the temporal features. Recently, 3D convolutional networks~\cite{ji20133d,carreira2017quo,feichtenhofer2018slowfast,wu2019long,girdhar2019video} appear, where the temporal information is treated as the same with the spatial ones. The large 3D kernel makes this method consume a large amount of computational resource.
		\paragraph{Methods to split temporal and spatial information}
		A simple method to split temporal-spatial information is to utilize relatively pure spatial information without temporal one to extract spatial features and pure temporal input for temporal ones. For example, two-stream models~\cite{simonyan2014two,carreira2017quo,feichtenhofer2016convolutional} adopt one static image as spatial input and optical flows as temporal input. One problem of this method is that the processes of extracting spatial and temporal features are completely independent, making it impossible to extract hierarchical spatial-temporal features. Another method is to adjust the density of these two types of information. In SlowFast network~\cite{feichtenhofer2018slowfast}, the input of spatial stream has higher spatial resolution and lower temporal sampling rate, while the input of temporal stream is the opposite.

	\section{Methods}~\label{Sec:Method}
	In this section, we will introduce the detailed structure of SCS, the training method with spatial-temporal switch gradient descent, the strategy to deal with the high-dimension spatial signal and super long sequences, and the designs of the experiments.
	
	\subsection{Network for SCS}
	At every time-stamp $t$, the network $\mathcal{T}$, consisting of $n$ semi-coupled layers, receives an input matrix $\mathbf{x}_t$ from the dataset or environment and outputs an vector $\mathbf{y}_t$ (the main goal $g$) to approximate the target (ground truth) vector $\mathbf{z}_t$.  
	
	As mentioned above, each semi-coupled layer satisfies the structure of $\mathbf{u}^{l}_t = \mathcal{F}(h_s(\mathbf{u}^{l-1}_t), h_t(\mathbf{u}^{l-1}_t))$, where $\mathbf{u}^{l}_t$ is the output of the $l^{th}$ layer at $t^{th}$ step and $\mathbf{u}^{l-1}_t$ is the input. By defining $\mathbf{u}^0_t=\mathbf{x}_t$, we get:  
	\begin{align} 
	& \mathbf{s}^l_t=  h_s(\mathbf{u}^{l-1}_t;\psi_s^l) = {\rm Conv}(\mathbf{u}^{l-1}_t; \psi_s^l) \label{eq:hs}\\
	& \mathbf{c}^l_t=  h_t(\mathbf{u}^{l-1}_t;\psi_t^l) = {\rm Conv}([\mathbf{u}^{l-1}_t, \sigma(\mathbf{c}^l_{t-1})];\psi_t^l) \label{eq:ht}
	\end{align}
	where $l$ is the layer index, $\sigma(x)=1/(1+exp(-x))$ is the logistic sigmoid function, $\rm Conv$ is the convolutional neural layer, $\psi_s^l$ and $\psi_t^l$ are spatial state and temporal cell state matrix, respectively, of layer $l$ at time $t$. $\mathbf{c}^l_0=\mathbf{0}$ is true for all $l$. We adopt $\rm Conv$ here for it is an excellent spatial feature extractor and of course, we can replace $\rm Conv$ by other operators like fully connection, according to different tasks. Note that Eq.~\ref{eq:hs} describes the structure of $h_s$ and Eq.~\ref{eq:ht} describes $h_t$ which is a simple naive RNN structure. It is feasible to replace $h_t$ with LSTM architecture~\cite{xingjian2015convolutional}, but the computing complexity is too high to apply on visual tasks, so we do not practice this in this paper.
	
	The synthesizer $\mathcal{F}$ adopts a parameter-free structure:
	\begin{align}
	& \mathbf{u}^l_t={\rm Relu}(\mathbf{s}^l_t) \circ {\rm Sigmoid}(\mathbf{c}^l_t) \label{eq:F_ts} 
	\end{align}
	
	where $\circ$ denotes element-wise multiplication, ${\rm Relu}(x)=max(0,x)$ is the rectified linear unit and ${\rm Sigmoid}(x)= 1/(1 + e ^ { - x })$ is the sigmoid function. In this way, the results of $h_t$ are normalized to $(0,1)$, so $h_t$ is treated as a control gate of $h_s$ in the viewpoint of $\mathcal{F}$.
	
	As the network is recurrent, its outputs are a function of the complete sequence $(\mathbf{x}_1,...,\mathbf{x}_t)$. We can further encapsulate the operation of the network as
	\begin{align}
	&(\mathbf{u}^n_1,..., \mathbf{u}^n_t)=\mathcal{T}([\mathbf{x}_1,...,\mathbf{x}_t];\Psi_s, \Psi_t)
	\end{align}
	where $\bm{\Psi}$ is the set of trainable network weights and $\mathbf{u}^n_t$ is the output of the $n$th layer at time stamp $t$. Finally, the output vector $\mathbf{y}_t$ is defined by the assembly of $(\mathbf{u}^n_1,..., \mathbf{u}^n_t)$:
	\begin{align}\label{eq:yt}
	& \mathbf{y}_t=[\mathbf{u}^n_1,..., \mathbf{u}^n_t]
	\end{align}
	
	For $\mathcal{T}^1$ and $\mathcal{T}^2$, the sub-goal networks, we adopt the same  $h_s(\cdot)$ and $h_t(\cdot)$ with $\mathcal{T}$, while the synthesizer $\mathcal{F}$ is different. In $\mathcal{T}^1$, $\mathcal{F}$ and sub-goal $r_s$ (or $\mathbf{y}_t^{\mathcal{T}^1}$) are defined as: 	
	\begin{align}
	\hat{\mathbf{u}}^l_t&={\rm Relu}(\mathbf{s}^l_t)\\
	\mathbf{y}_t^{\mathcal{T}^1}&=[\hat{\mathbf{u}}^n_1,..., \hat{\mathbf{u}}^n_t]
	\end{align}
	While in $\mathcal{T}^2$, $\mathcal{F}$ and sub-goal $r_t$ (or $\mathbf{y}_t^{\mathcal{T}^2}$) are defined as: 	
	\begin{align}
	\hat{\mathbf{u}}^l_t&={\rm Relu}(\mathbf{c}^l_t)\\
	\mathbf{y}_t^{\mathcal{T}^2}&=[\hat{\mathbf{u}}^n_1,..., \hat{\mathbf{u}}^n_t]
	\end{align}
	\subsection{Deep Nested Semi-coupled Structure Training}~\label{sec:STSGD}
	
\begin{figure}
		\begin{center}
			\includegraphics[width=\linewidth]{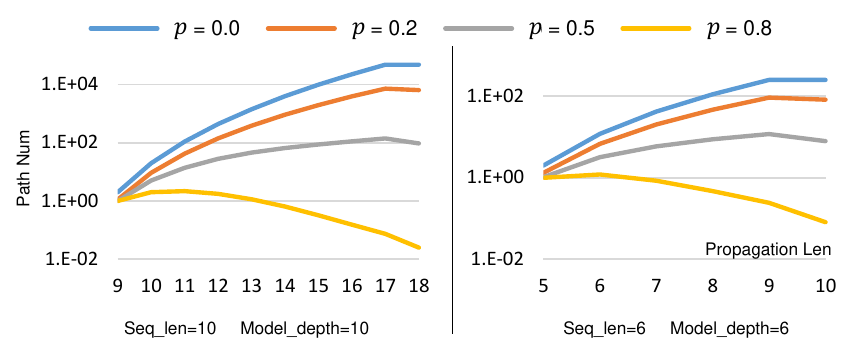}
		\end{center}
		\caption{\textbf{Expectation numbers of back-propagation chains}. The horizontal axis is the length of the back-propagation chain and the vertical axis is the exception number of the chains. Note that with the growing of the model depth and sequence length, the number and length of the chains grow significantly. Our STSGD with large $p$ can efficiently reduce the number of long sequences.}
		\label{fig:STSGD} 
	\end{figure}

	As discussed in section \ref{sec:aware}, on one hand, $\mathcal{G}$ computes spatial and temporal information by separate modules. On the other hand, we adopt deep nested structure of stacked $\mathcal{G}$, inspired from the spatial and temporal coupling in human brains. But this structure leads to a difficult training process, because the deep nested structure actually merges the spatial and temporal information early in the shallow layers, which aggravates the pressure of spatial and temporal decomposition in the later layers as well as reduces the hierarchy of the decoupled features. Moreover, as the depth of layers and the length of sequences increase, both the number and the length of the back-propagation chains will grow significantly, which makes the training process much more challenging as well (see Fig.~\ref{fig:STSGD}).

	To address this challenge, the Spatial-Temporal Switch Gradient Descent (STSGD) is proposed to conduct a higher level of semi-coupling, in which the optimizer updates parameters based on either spatial or temporal information with a certain probability at each training step.
	As the training goes on, we reduce the degree of this separation and finally the network can learn all the information. This training strategy is also a practice of the semi-coupled mechanism: decoupling first then synthesizing.

	\paragraph{Spatial-Temporal Switch Gradient Descent}
	
	STSGD is also a gradient based optimization method and the gradients are propagated by the BP algorithm~\cite{rumelhart1988learning}. It works like a switch that turns off gradients on spatial and temporal modules with a certain probability. This scheme largely reduces the interference between $h_s(\cdot)$ and $h_t(\cdot)$ induced by the deep nested structure.
	
	Given the definition Eq.~\ref{eq:G}. Its forward propagation is:
	\begin{align}
	& \mathbf{y}_t = \mathcal{G}_n \circ \ldots \circ \mathcal{G}_1 
	\end{align}
	where $\mathcal{G}_i = \mathcal{F}[h_t(\cdot), h_s(\cdot);\psi_i]$ is the $i^{th}$ layer of the network and $\bm{\psi}_i$ is the set of the trainable parameters in the $i^{th}$ layer. The loss between $\mathbf{y}_t$ and ground truth $\mathbf{z}_t$ is defined as:
	\begin{align}
	E = \sum_{t=1}^T{E_t} = \sum_{t=1}^T{L(\mathbf{y}_t, \mathbf{z}_t)}
	\end{align}
	where $L$ is the loss function.
	
	Then during back-propagation, according to the BPTT argorithm, the gradient of $\bm{\psi}_i$ can be presented as:
	\begin{equation}
	\begin{aligned}
	\frac{\partial E}{\partial \bm{\psi}_i} = & \sum_{t=1}^T\frac{\partial E}{\partial \mathcal{G}^t_{i+1}}\frac{\partial\mathcal{G}^t_{i+1}}{\partial \bm{\psi_i}}
	\end{aligned}
	\end{equation}
	and in conventional stochastic gradient descent methods, this exact gradient is adopted to update the parameters and continue back propagating. In our STSGD, we need to decouple the gradients based on the information carried by these two modules. To this end, we rewrite the gradient as:
	\begin{equation}
	\begin{aligned}
	\frac{\partial E}{\partial \bm{\psi}_i} & = \sum_{t=1}^T\frac{\partial E}{\partial \mathcal{G}^t_{i+1}}(\frac{\partial \mathcal{G}^t_{i+1}}{\partial \mathbf{c}^{i+1}_{t}}\frac{\partial \mathbf{c}^{i+1}_{t}}{\partial \bm{\psi}_i} + \frac{\partial \mathcal{G}^t_{i+1}}{\partial \mathbf{s}^{i+1}_{t}}\frac{\partial \mathbf{s}^{i+1}_{t}}{\partial \bm{\psi}_i})
	\end{aligned}
	\end{equation}
	
	In the equation, the first term in the bracket is the gradient from $h_t(\cdot)$ and the second term is from $h_s(\cdot)$. As the structures of $h_t(\cdot)$ and $h_s(\cdot)$ are designed for temporal and spatial information respectively, the gradients from them carry different concepts.
	
	To decouple the gradients, we use a switch to prevent a certain part (spatial or temporal) from propagating its gradient in back-propagation, which can be defined as:
	\begin{equation}
	\begin{aligned}
	\hat{\frac{\partial E}{\partial \bm{\psi}_i}} =\sum_{t=1}^T(&\gamma_t(p_t)\frac{\partial E}{\partial \mathcal{G}^t_{i+1}}\frac{\partial \mathcal{G}^t_{i+1}}{\partial \mathbf{c}^{i+1}_{t}}\frac{\partial \mathbf{c}^{i+1}_{t}}{\partial \bm{\psi}_i}\\
	+&
	\gamma_t(p_s)\frac{\partial E}{\partial \mathcal{G}^t_{i+1}}\frac{\partial \mathcal{G}^t_{i+1}}{\partial \mathbf{s}^{i+1}_{t}}\frac{\partial \mathbf{s}^{i+1}_{t}}{\partial \bm{\psi}_i}) \label{eq: split}
	\end{aligned}
	\end{equation}
	where $\gamma$ is a probability function defined as:
	\begin{equation}
	\gamma(p) =\left\{
	\begin{aligned}
	0, &{\rm~with~the~probability~of~}p\\
	1, &{\rm~with~the~probability~of~}(1-p)
	\end{aligned}
	\right.
	\end{equation}
	
	\paragraph{Discussion} This scheme partly decouples the spatial and temporal learning process by initializing $p_s$ and $p_t$ to a relative high value ($p_s=p_t=0.5$) and as the training goes on, $p$ decreases to 0 to synthesize the spatial and temporal training processes. From a macro perspective, it cuts off some paths in the back propagation with a certain probability, which reduces the number of back-propagation chains significantly, to make the training process more tractable (see Fig.~\ref{fig:STSGD}). According to the Assumption 4.3 in~\cite{bottou2018optimization}, if we set $p_s = p_t$, we get $E(\hat{\frac{\partial E}{\partial \bm{\psi}_i}})={\frac{\partial E}{\partial \bm{\psi}_i}}$ and this will lead to the similar convergence properties with the conventional stochastic gradient descent method. 
	
	\paragraph{Advanced STSGD}
	Note that, in STSGD, the same value of $p$ for $h_s(\cdot)$ and $h_t(\cdot)$ is the restriction for convergence and is a sufficient condition. But, we hope the network can learn more spatial information at the beginning, since temporal information can not be captured given a very unreliable spatial representation.
	After getting a relatively mature spatial representation, we hope the STSGD can shift its learning focus between spatial and temporal features. To this end, we modify the Eq.~\ref{eq: split} with a dynamic ratio $q\in [0,1]$ to: 
	\begin{equation}
	\begin{aligned}
	\hat{\frac{\partial E}{\partial \bm{\psi}_i}} =\sum_{t=1}^T(&\gamma_t(q)\frac{\partial E}{\partial \mathcal{G}^t_{i+1}}\frac{\partial \mathcal{G}^t_{i+1}}{\partial \mathbf{c}^{i+1}_{t}}\frac{\partial \mathbf{c}^{i+1}_{t}}{\partial \bm{\psi}_i}\\
	+&
	\gamma_t(1-q)\frac{\partial E}{\partial \mathcal{G}^t_{i+1}}\frac{\partial \mathcal{G}^t_{i+1}}{\partial \mathbf{s}^{i+1}_{t}}\frac{\partial \mathbf{s}^{i+1}_{t}}{\partial \bm{\psi}_i}) \label{eq:split2}
	\end{aligned}
	\end{equation}

	Although there is no theory to guarantee the convergence of the Advanced STSGD (ASTSGD), the experiment results show it works well. Moreover, to automatically control the process of decreasing $q$, we train a small network with 3 fully connection layers which takes $r_s$, $r_t$, and $g$ as input to optimize $q$. To simplify the problem, we provide an empirical formula as another option:
	\begin{align}
	q =q_0 + (1-q_0)\frac{\max(0, L_s- {\rm thresh})}{{\rm InitL_g} - {\rm thresh}} * (\alpha(\frac{L_g}{L_s} -1)+1)
	\end{align}
	where $L_s$ and $L_g$ are the loss values of $r_s$ and $g$. $q_0$ is usually set as 0.5. In this equation, we monitor the decreasing process of $L_s$ to update $q$. $\rm thresh$ is a hyper-parameter that acts as a threshold for $L_s$, considering that $L_s$ is difficult to decrease to 0 and we want $q$ get the minimum value when $L_s$ decreasing to $\rm thresh$. $\rm InitL_g$ is the initial training loss value of $g$, for example, the initial loss of an $n$-class classification problem, the initial cross entropy loss, is $ln(n)$. $\alpha$ is a hyper-parameter and the multiplier $(\alpha(L_g/L_s-1)+1)$ is designed to balance the integral and spatial information. If the task is more depended on the spatial information, we can set $\alpha$ smaller, in this way the function will have a relative big value to better learn the spatial features.
	
	\subsection{Dealing long sequences with LTSC}\label{sec:LTSC}
	Making use of the different properties between time and space, $\mathcal{T}[\mathbf{x};\Psi_s, \Psi_t]$ decouples the temporal and spatial information. As the length of the sequence grows, the enormous increase in information which leads to huge computing requirements also brings up the demand of information decoupling. Briefly, \textit{long range temporal semi-coupling} (LTSC) decouples the original long sequence $\mathbf{D}$ into short sequences $\{\mathbf{d}_i|i = 1,...,n\}$ along the temporal dimension with a partitioning principle:
	\begin{align}
	& \bigcup_{i=1,...,n}\{\tilde{\mathbf{d}_i}\} =\tilde{\mathbf{D}} \\
	& \tilde{\mathbf{d}}_{i-1} \cap \tilde{\mathbf{d}}_i \neq \varnothing \label{eq: overlap}
	\end{align}
	where $\tilde{\mathbf{d}}_i$ and $\tilde{\mathbf{D}}$ are the set of $\mathbf{x}_i$ contained in sequence $\mathbf{d}_i$ and $\mathbf{D}$, and ${\mathbf{d}_i}$ is sorted by index of its first element. Eq.~\ref{eq: overlap} requires overlaps between the adjacent sub-sequences which are the hinge to transmit the information, for it makes sure that there is no such a point in $\mathbf{D}$ that information cannot feed forward and backward across it. For example, when splitting $\mathbf{D} = (\mathbf{x}_1, \mathbf{x}_2,\mathbf{x}_3, \mathbf{x}_4)$ into $\mathbf{d}_1=(\mathbf{x}_1, \mathbf{x}_2)$ and $\mathbf{d}_2=(\mathbf{x}_3, \mathbf{x}_4)$ in which there is no overlap, information will never transfer between $\mathbf{x}_2$ and $\mathbf{x}_3$, but if splitting $\mathbf{D}$ into $\mathbf{d}_1 = (\mathbf{x}_1, \mathbf{x}_2, \mathbf{x}_3)$ and $\mathbf{d}_2=(\mathbf{x}_2, \mathbf{x}_3, \mathbf{x}_4)$, there is no such issue.
	
	As a hyper-parameter, a high overlap $\eta$ will lead to high computing complexity and a low $\eta$ means worse ability to transmit information. In this paper, we chose 25\% for $\eta$.
	
	LTSC can be adopted in any sequence task, while in this paper, LTSC is nested with SCS network.
	LTSC further utilizes the overlaps among $\mathbf{d}_i$ to enhance the flow of information and we adopt an error function to shorten the distance between output of the adjacent $\mathbf{d}_i$:
	\begin{align}
	& L_{overlap}=\sum_{i=2,...,n}{\xi(h_m({\mathbf{d}}_{i-1}), h_m({\mathbf{d}}_i))}
	\end{align}
	where $\xi(a, b)$ is defined as the overlap MSE function which calculates the MSE value of the overlap parts of $a$ and $b$. This arrangement makes the output of $h_m$ with adjacent input $\mathbf{d}_i$ be close in the overlap part.
	
	Note that there are other straightforward methods to decouple the long-range temporal information like TBPTT algorithm~\cite{williams1990efficient} or simply sampling from the original $\mathbf{D}$~\cite{carreira2017quo,donahue2015long,gu2017ava,hou2017tube}, but these do not make sure that the semantic information can transmit through the whole sequence or just discard some percentage of information.
	
	A simple example demonstrates the smooth flow of semantic information in LTSC. The input visual sequence consists of an image of star and several of rhombus, and the star can appear at any temporal position. Our model needs to learn how far the current rhombus image is from the appeared star. With LTSC, the model can correctly output the results even if the star image appears 50 frames ago when the decoupled sequence lengths are smaller than 10. This can serve as a preliminary verification on LTSC.
	
	\begin{figure*}
		\begin{center}
			\includegraphics[width=\linewidth]{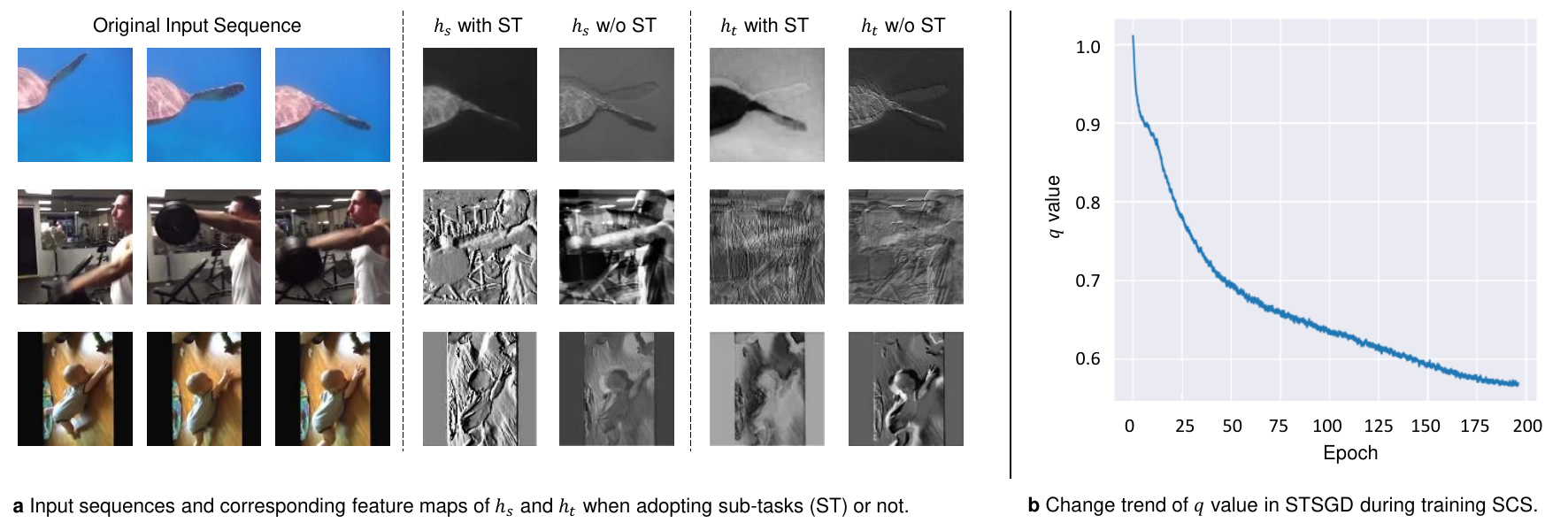}
		\end{center}
		\caption{\textbf{Ablation study on action recognition task}. \textbf{a}, Feature maps of $h_s$ and $h_t$. With sub-tasks, the splitting of spatial and temporal information is more obvious than without sub-tasks: 1) With sub-tasks, $h_s$ contains purer spatial information, while without sub-tasks, there is a little temporal information in $h_s$. 2) Without sub-tasks, $h_t$ also extracts texture information, while sub-tasks makes it focus on the motion changes. This reveals that sub-tasks can make $h_s$ and $h_t$ focus on their own jobs. \textbf{b}, $q$ values during the training process. At the beginning of training, $q$ is about $1$, which leads SCS to focus on spatial information first. As the training goes on, the value of $q$ gradually approaches $0.5$ to merge temporal and spatial information and the model treats them equally.}
		\label{fig:ablation_study}
	\end{figure*}
	
	\begin{figure*}
		\begin{center}
			\includegraphics[width=\linewidth]{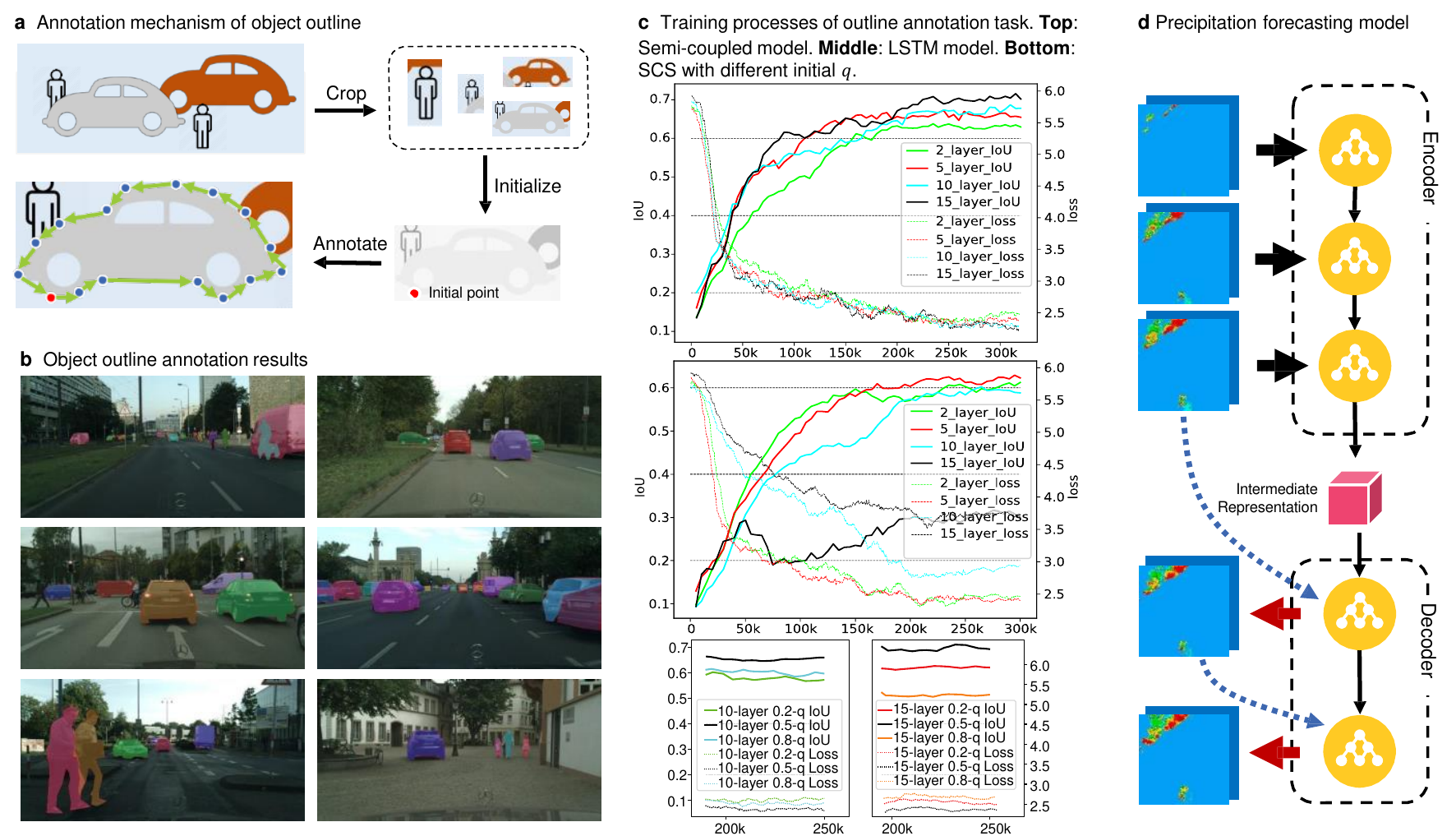}
		\end{center}
		\caption{\textbf{Outline annotation and precipitation forecasting experiments}. \textbf{a}, To annotate the oulines, we first crop out the objects, then give the start point (the red one), finally the model needs to give out the outline points starting from it one by one to form a complete outline. \textbf{b}, Some example images annotated by our SCS structure. \textbf{c}, From the training process of the model on outline annotation, we can see that our SCS model can benefit from the increasing model depth while LSTM model is difficult to train when stacked deep, which is because extracting the hierarchical spatial and temporal features together is difficult and our Semi-Coupled Structure can solve this problem. Moreover, a larger initial $q$ can better decouple the training process of $h_s$ and $h_t$, thus leads to a better performance. \textbf{d}, The precipitation forecasting model consists of an encoder and a decoder. The encoder integrates the observed radar echo images into an intermediate representation. The decoder takes the representation and the last radar echo image as input, updates the intermediate representation and outputs the future image one by one. }
		\label{fig:experiments}
	\end{figure*}
	
	\subsection{Comparison with Deep RNN and spatial-temporal attention model}
	
	The Deep RNN framework~\cite{pang2018deep} is the predecessor to the SCS described in this work, yet they have significant differences. 
	Firstly, in the Deep RNN framework, the splitting of two flows is designed to make the deep recurrent structure easier to train by adding spatial shortcuts over temporal flows. While in SCS, the semi-coupling mechanism aims at endowing the model with the awareness of spatial and temporal concepts. Moreover, $h_s(\cdot)$ and $h_t(\cdot)$ have the equal status to explicitly learn the two concepts. 
	Secondly, the Deep RNN framework has no mechanism to ensure the two flows focusing on the two kinds of information. This is not an issue for the SCS which adopts two extra independent sub-goals and two stand-alone modules tailored for spatial and temporal features. Thirdly, in the training process, Deep RNN has no way to control the training degree of the two flows, thus, no way of re-focusing on the spatial or temporal information. This problem is addressed in SCS by the STSGD mechanism.

	The Spatial-Temporal Attention model (STAM)~\cite{song2017end} also introduces spatial and temporal concepts. There are several differences between STAM and SCS. Firstly, SCS aims at extracting the temporal and spatial features (concepts) separately from the input, while STAM is designed to output the spatial and temporal attention from input features which integrate spatial and temporal information. These attentions defined on skeleton key-points rely on human skeleton assumption very much and are not general features. Secondly, STAM is designed for small-scale data format (skeleton coordinates, 20D only), thus, it is not suitable for the large scale problems which are the targets of SCS. Thirdly, STAM does not have awareness of general spatial and temporal concepts. The heads of spatial and temporal attention modules are designed for the specific tasks: spatial one for skeleton keypoints and temporal one for video frames, thus, the spatial and temporal concepts are actually human-defined, not learnt by the model itself.

	\subsection{Action recognition task descriptions}
	The experiments of action recognition are conducted on UCF-101, HMDB-51 and Kinetics-400 datasets, which comprise sets of 101, 51, 400 action categories respectively. For each dataset, we follow the official training and testing splits. For each video, the frames are rescaled with their shorter side into 368 and a 224 $\times$ 224 crop is randomly sampled from the rescaled frames or their horizontal flips. Colour augmentation is used, where all the random augmentation parameters are shared among the frames of each video.
	
	For this task, we adopt two kinds of Semi-Coupled Structures: backbone-supported one and stand-alone structure without backbone. For the backbone-supported structure, there is a CNN backbone pre-trained on ImageNet~\cite{krizhevsky2012imagenet} (we choose VGG and InceptionV1 as examples) before the 15-layer SCS network. While for the stand-alone version, the model only consists of a 17-layer SCS network. Shortcuts between layers like ResNet~\cite{he2016deep} are adopted in SCS networks to simplify the training process. The detailed structures are summarized in Tab.~\ref{tab:actionstructure}.
	
	\begin{table}[]
		\footnotesize
		\caption{Detailed stand-alone SCS structures for action recognition task. There are residual lines between every layer blocks.
		}
		\renewcommand{\arraystretch}{2}
		\begin{center}
			\begin{tabular}{c|c}
				\hline
				layer blocks &  output size\\
				\hline
				$7\times7, 64, {\rm stride}2$ & 112\\[1.5ex]
				$ \left[ \begin{aligned}
				3\times3, 64 \\
				3\times3, 64
				\end{aligned}
				\right] \times 2$& 56\\[4ex]
				$ \left[ \begin{aligned}
				3\times3, 128 \\
				3\times3, 128
				\end{aligned}
				\right] \times 2$& 28\\[4ex]
				$ \left[ \begin{aligned}
				3\times3, 256 \\
				3\times3, 256
				\end{aligned}
				\right] \times 2$& 14\\[4ex]
				$ \left[ \begin{aligned}
				3\times3, 512 \\
				3\times3, 512
				\end{aligned}
				\right] \times 2$& 7\\
			\end{tabular}
		\end{center}
		\label{tab:actionstructure}
		\vspace{-0.2in}
	\end{table}

	The main goal $g$ of the network is to minimize the cross-entropy of the softmax outputs with respect to the action categories; the final output is the average of the outputs of every time-stamp frame. The spatial goal $r_s$ is the same as the main goal and the temporal goal $r_t$ is to estimate the optical flow between the current and last input frames. For each step, the network processes a new video frame and the probability distribution over action categories is predicted based on the current processed frames.
	
	Adopting LTSC enables our network to process much longer sequences than previous works on action recognition in which sampling methods are used to shorten the video length. This places greater stress on the long-range memory capacity of the model but preserves more temporal information in the original video. In addition, due to the deep structure of SCS network, we adopt ASTSGD.
	
	Tab.~\ref{tab:actionResult} lists the complete results and hyper-parameters of the experiments on action recognition for SCS, LSTM and pure CNN model. We can see that our SCS has much better performances than LSTM, ConvLSTM, and CBM~\cite{pang2018deep} models. Compared with CBM, the new SCS decouples spatial and temporal information and adjusts its focus (on spatial or temporal information) strategically during the learning process. Detailed analysis is shown in ``Ablation Study".
	\begin{table}[]
		\footnotesize
		\caption{Action recognition accuracy on Kinetics, and end-to-end fine-tuning on UCF-101 and HMDB-51. Note that our SCS model applies 17 layers. ``BB" denotes backbone. 
		}
		\renewcommand{\arraystretch}{1.2}
		\begin{center}
			\begin{tabular}{c|c|c|c}
				\hline
				Architecture &  Kinetics & UCF-101 & HMDB-51\\
				\hline
				\multicolumn{2}{c|}{}& \multicolumn{2}{c}{Pre-trained on Kinetics}\\
				\hline
				LSTM with BB (VGG)~\cite{donahue2015long} & 53.9 & 86.8 & 49.7\\
				3D-Fused~\cite{feichtenhofer2016convolutional} & 62.3 & 91.5 & 66.5\\
				Stand-alone CBM~\cite{pang2018deep} & 60.2 & 91.9  & 61.7 \\
				Stand-alone SCS & 61.7 & 92.6  & 65.0 \\
				\hline
				\multicolumn{2}{c|}{}& \multicolumn{2}{c}{Not pre-trained on Kinetics}\\
				\hline
				15-layer ConvLSTM & - & 68.9 & 34.2\\
				BB (VGG) supported CBM~\cite{pang2018deep} &-& 79.8 & 40.2\\
				BB (VGG) supported SCS &-& 82.1 & 42.5\\
				BB (Inception) supported SCS &-& 87.9 & 52.1\\
				
				\hline
			\end{tabular}
		\end{center}
		\label{tab:actionResult}
		\vspace{-0.2in}
	\end{table}
	
	\paragraph{Ablation Study}
	Since our SCS is a universal backbone, we conduct the ablation study on this low-level feature-driven task to show the function of each component. The results are shown in Tab.~\ref{tab:ablsty}. We first test the design of the spatial-temporal sub-task paradigm. From the view of the performances, it leads to 1.2\% accuracy boost and from Fig.~\ref{fig:ablation_study} \textbf{a}, we can see that this paradigm makes $h_s$ and $h_t$ more focus on their own functions: $h_s$ for spatial features and $h_t$ for temporal ones. Then we remove the ASTSGD from the training process, leading to 1.4\% accuracy drop. In Fig.~\ref{fig:ablation_study} \textbf{b}, we show the change tendency of $q$, which demonstrates that the model learns spatial information first then merges temporal features into it just as we expect. The sub-tasks and ASTSGD are the main improvements on CBM~\cite{pang2018deep}, which make $h_s$ and $h_t$ focus on their jobs, the training process controllable, and the model perform better. Without LTSC, the model can only access a short clip due to the limit of computational resource and the accuracy drops 2\%.

	\begin{table}[]
		\footnotesize
		\caption{Ablation study results (accuracy) on action recognition task with Kinetics and UCF-101 dataset. ``w/o" denotes ``without".}
		\renewcommand{\arraystretch}{1.2}
		\begin{center}
			\begin{tabular}{c|c|c}
				\hline
				Architecture &  Kinetics & HMDB-51\\
				\hline
				Whole Stand-alone SCS & 61.7 & 65.0\\
				Stand-alone SCS w/o two sub-tasks & 60.5 & 63.7\\
				Stand-alone SCS w/o sub-task $T^2$ only & 61.3 & 64.2\\
				Stand-alone SCS w/o ASTSGD & 60.8 & 63.2\\
				Stand-alone SCS w/o LTSC & 59.7 & 62.9\\
				\hline
			\end{tabular}
		\end{center}
		\label{tab:ablsty}
		\vspace{-0.2in}
	\end{table}
	
	\subsection{Outline annotation task descriptions}
	We adopt CityScapes dataset~\cite{cordts2016cityscapes} to conduct the outline annotation task experiments. CityScapes dataset consists of the street view images and their segmentation labels. We crop out the backgroud of the images and only preserve 8 kinds of the foregrounds: Bicycle, Bus, Person, Train, Truck, Motorcycle, Car and Rider. After cropping, there are 51k training images and 10k test images. The preserved images are resized to 224 $\times$ 224.
	
	Similar with Polygon-RNN~\cite{castrejon2017annotating,acuna2018efficient}, a VGG model is adopted first to extract the spatial features of the original images. Then a deep SCS network with 15 layers takes the image features and the outline point positions of time-stamp $t-1$ as input for each time-stamp $t$ and generates the outline point positions sequentially. In short, we replace the RNN part in the
	original Polygon-RNN model with our deep SCS network and adjust the optimizing method with our LTSC and STSGD schemes.
	
	This task is also treated as a classification task. Each position of the image is a class and the loss function is the cross-entropy of the softmax outputs with respect to the image positions (28 $\times$ 28 $+$ 1, 784 positions in total and a terminator). The spatial goal $r_s$ and temporal goal $r_t$ are the same as the main goal: predicting the positions of the outline's key points. Though we do not adopt different targets for $\mathcal{T}^1$ and $\mathcal{T}^2$, the independent optimization processes and asymmetrical structures allow them to focus on different information. The outline position sequence makes up a polygon area and we adopt the IoU between the predicted and ground-truth polygon area as the evaluation metric.
	
	For each foreground, the model predicts 60 outline points at most. With a 15-layer SCS network, this sequence length requires huge computing resources, so we adopt the LTSC mechanism to split the sequence into 6 short clips and utilize ASTSGD to decouple the temporal and spatial training process of the deep structure.
	
	Detailed results and hyper-parameters of the experiments on outline annotation for SCS-Polygon-RNN, Polygon-RNN and Polygon-RNN++ are shown in Tab.~\ref{tab:polygonResult}. Compared with traditional LSTM or GRU module, our model can be stacked deeply and achieve better performances with less parameters. Our SCS does not achieve the state-of-the-art performance because Polygon-RNN++ adopts many advanced tricks to improve the performance (including reinforcement learning, graph neural network, and attention module). These tricks are not the focus of this paper. Compared with CBM~\cite{pang2018deep}, the new SCS with sub-tasks and ASTSGD achieves better performances.
	\begin{table}[]
		
		\caption{Performance (IoU in \%) on Cityscapes validation set (used as test set in~\cite{castrejon2017annotating}). Note that ``Polyg-LSTM" denotes the original Polygon-RNN structure with ConvLSTM cell, ``Poly-GRU" for Polygon-RNN with GRU cell, and ``Polyg-SCS" for Polygon-RNN with our Semi-Coupled Structure.}
		\renewcommand{\arraystretch}{1.2}
		\begin{center}
			\footnotesize
			\begin{tabular}{c|c|c|c}
				\hline
				\multicolumn{3}{c|}{Model} & IoU\\
				\hline
				\multicolumn{3}{c|}{Original Polygon-RNN~\cite{castrejon2017annotating}} & 61.4\\
				\multicolumn{3}{c|}{Residual Polygon-RNN~\cite{acuna2018efficient}} & 62.2\\
				\multicolumn{3}{c|}{Residual Polygon-RNN + attention + RL~\cite{acuna2018efficient}} & 67.2\\
				\multicolumn{3}{c|}{Residual Polygon-RNN + attention + RL + EN~\cite{acuna2018efficient}} & 70.2\\
				\multicolumn{3}{c|}{Polygon-RNN++~\cite{acuna2018efficient}} & \textbf{71.4}\\
				\hline
				& \# layers & \# params of RNN & \\
				\hline
				Polyg-LSTM & 2 & 0.47M & 61.4\\
				Polyg-LSTM & 5 & 2.94M & 63.0\\
				Polyg-LSTM & 10 & 7.07M & 59.3\\
				Polyg-LSTM & 15 & 15.71M & 46.7\\
				\hline
				Polyg-GRU & 5 & 2.20M & 63.8\\
				Polyg-GRU & 15 & 11.78M & 64.7 \\
				\hline
				Polyg-CBM~\cite{pang2018deep} & 5 & 1.13M & 63.1 \\
				Polyg-CBM~\cite{pang2018deep} & 15 & 5.85M & 70.4\\
				\hline
				Polyg-SCS & 2 & 0.20M & 62.9\\
				Polyg-SCS & 5 & 1.13M & 65.8 \\
				Polyg-SCS & 10 & 2.68M & 68.0\\
				Polyg-SCS & 15 & 5.85M & \textbf{71.0}\\
				\hline
			\end{tabular}
		\end{center}
		\label{tab:polygonResult}
		\vspace{-0.2in}
	\end{table}
	
	\subsection{Auto-driving task description}
	
	Auto driving is a complex task. Completely solving it requires to conduct scene sensing, route planning, security assurance and so on. Here we simplify the task into a sequential vision task: given a short video from driver's perspective and outputting the driving direction in the form of steering wheel angles. The experiments are conducted on comma.ai~\cite{santana2016learning} and LiVi-Set~\cite{chen2018lidar} datasets. These sets record the real driving behaviours of human drivers and there are various road conditions including town streets, highways and mountain roads. To make the model better focus on the road, we crop out the sky and other irrelevant information from the original images. And the final input images are resized to 192 $\times$ 64.
	
	We compare our SCS network with conventional LSTM model and CNN model. The SCS structure is the same as the stand-alone model used in action recognition and the LSTM model adopts a CNN backbone (VGG) like LRCNs~\cite{donahue2015long}. These two models both take a short driving video as the input and extract the temporal-spatial features. While for CNN model, we adopt the ResNet~\cite{he2016deep} structure, and it only takes the current driving image as input and utilizes spatial features to commit predicting. 
	
	This is a regression problem and the main goal $g$ of the network is set to minimize the MSE of the predicted steering angles with the ground truth. The same with the action recognition task, the spatial goal $r_s$ is the same as the main goal and the temporal goal $r_t$ is to estimate the optical flow. We adopt sigmoid function to normalize the angles because the angles before normalization are more likely distributed around 0 and this non-linear function can, to some extent, make the distribution more uniform. Accuracy is used as the metric which is defined as:
	\begin{align}
	Acc=\frac{\sum_i^I\lfloor{\min(\frac{\lambda}{|{\rm pred}_i-{\rm label}_i| + \epsilon }, 1)}\rfloor}{I}
	\end{align}
	where $I$ is the number of the samples, $\lambda$ is a threshold, and $\epsilon$ is a small value to prevent the denominator from being zero. ${\rm pred}_i$ and ${\rm label}_i$ are the predicted angle value and label angle value of sample $i$. In short, if the difference between predicted angle and label angle is less than the threshold, we treat it as an accurate prediction.
	
	To predict the current driving direction, the models need to review a short history driving video (except the CNN model). We adopt our LTSC scheme when reviewing relative long history and we adopt STSGD for the deep SCS networks. We find that, adopting LTSC to access more temporal information makes the model achieve better performances without increasing memory resources. And STSGD relatively improves the performances by 9\% on average. The detailed comparison results are shown in Tab.~\ref{tab:drivingResult}.
	
	\begin{table}[]
		
		\caption{Auto-driving performance of SCS and baselines (CNN, CNN+LSTM) on the comma.ai and LiVi-Set validation set. Note that ``$\lambda$" denotes the angle threshold, ``p" denotes the initial probability to stop the back-propagation in STSGD and ``length" denotes the number of observed frames.}
		\renewcommand{\arraystretch}{1.2}
		\begin{center}
			\footnotesize
			\begin{tabular}{c|c|c|c|c|c|c|c}
				\hline
				\hline
				\multicolumn{8}{c}{SCS model} \\
				\hline
				\multicolumn{2}{c|}{} & \multicolumn{3}{c|}{length=7} & \multicolumn{3}{c}{length=3}\\
				\cline{3-8}
				\multicolumn{2}{c|}{} & $\lambda$=6 & $\lambda$=3 & MSE & $\lambda$=6 & $\lambda$=3 & MSE\\
				\hline
				\multirow{3}{*}{LiVi} & p=0.0 & 31.8 & 16.9 & 0.046 & 28.8 & 15.9 & 0.049\\
				& p=0.3 & 34.1 & 17.4 & 0.045 & 30.5 & 16.1 & 0.048\\
				& p=0.5 & \textbf{35.1} & \textbf{19.4} & \textbf{0.044} & \textbf{33.4} & \textbf{17.6} & \textbf{0.046}\\
				\hline
				\multirow{3}{*}{Comma} & p=0.0 & 45.4 & 24.5 & 0.060 & 42.5 & 22.9 & 0.05 \\
				& p=0.3 & 48.8 & \textbf{25.5} & 0.043 & 46.9 & 23.9 & 0.044\\
				& p=0.5 & \textbf{49.2} & 25.0 & \textbf{0.037} & \textbf{47.4} & \textbf{24.1} & \textbf{0.041}\\
				\hline
				\hline
				\multicolumn{8}{c}{CNN+LSTM} \\
				\hline
				\multicolumn{2}{c|}{} & \multicolumn{3}{c|}{length=7} & \multicolumn{3}{c}{length=3}\\
				\cline{3-8}
				\multicolumn{2}{c|}{} & $\lambda$=6 & $\lambda$=3 & MSE & $\lambda$=6 & $\lambda$=3 & MSE\\
				\hline
				\multicolumn{2}{c|}{LiVi} & 29.2 & 15.9 & 0.052 & 27.3 & 14.5 & 0.057\\
				\hline
				\multicolumn{2}{c|}{Comma} & 43.1 & 23.8 & 0.056 & 42.3 & 21.0 & 0.058\\
				\hline
				\hline
				\multicolumn{8}{c}{CNN} \\
				\hline
				\multicolumn{2}{c|}{} & \multicolumn{3}{c|}{length=1} & \multicolumn{3}{c}{}\\
				\cline{3-5}
				\multicolumn{2}{c|}{} & $\lambda$=6 & $\lambda$=3 & MSE & \multicolumn{3}{c}{}\\
				\cline{1-5}
				\multicolumn{2}{c|}{LiVi} & 24.5 & 13.0 & 0.057 &\multicolumn{3}{c}{}\\
				\multicolumn{2}{c|}{Comma} & 45.2 & 25.3 & 0.056 &\multicolumn{3}{c}{}\\
				\hline

			\end{tabular}
		\end{center}
		\label{tab:drivingResult}
		\vspace{-0.2in}
	\end{table}
	
	We adopt this experiment to show that our SCS can quantitatively indicate how important the temporal information is toward the final goal. In this analysis, we set $r_s$ and $r_t$ to the same with the main goal. By calculating the accuracy of $\mathcal{T}^1$ and $\mathcal{T}^2$, we can determine the importance of temporal and spatial information in different road conditions. The results shown in Tab.~\ref{tab:tsimp} are consistent with our intuition: On straight roads, $\mathcal{T}^1$ and $\mathcal{T}^2$ have similar performances, which reveals that the temporal information is not so important on this condition. While on crossroads, $\mathcal{T}^2$ performs much better than $\mathcal{T}^1$, which shows that we need more temporal information to give out steering angles. On these four conditions, the performance gaps of $\mathcal{T}^2$ and $\mathcal{T}^1$ can be ordered as: straight roads \textless cure roads \textless T-junctions \textless crossroads. This is reasonable and proves that $\mathcal{T}^2$ and $\mathcal{T}^1$ focus on temporal and spatial information separately. Moreover, it preliminarily shows that how can we utilize this method to reveal the importance of temporal information on a specific sample.
	
	\begin{table}[]
		
		\caption{Accuracy of $\mathcal{T}^1$ and $\mathcal{T}^2$ on LiVi. Comparing their performances, we can get the importance of temporal information on different road conditions.}
		\renewcommand{\arraystretch}{1.2}
		\begin{center}
			\footnotesize
			\begin{tabular}{c|c|c|c|c|c|c|c}
				\hline
				\hline
				\multicolumn{2}{c|}{} & \multicolumn{2}{c|}{$\mathcal{T}^1$} & \multicolumn{2}{c|}{$\mathcal{T}^2$} & \multicolumn{2}{c}{$\mathcal{T}^2 - \mathcal{T}^1$}\\
				\cline{3-8}
				\multicolumn{2}{c|}{} & $\lambda$=6 & $\lambda$=3 & $\lambda$=6 & $\lambda$=3 & $\lambda$=6 & $\lambda$=3 \\
				\hline
				\multirow{4}{*}{\rotatebox{90}{LiVi}} & Crossroads &18.3 & 10.2 & 29.1 & 16.3 & 10.8 & 6.1\\
				& T-junction & 23.4 & 12.6 &32.2 & 17.0 & 8.8 & 4.4\\
				&  Curve road & 32.9 & 17.1 & 37.6 & 20.1 & 4.7 & 3.0\\
				&  Straight road & 39.1 & 21.0 & 41.7 &21.4 & 2.6 & 0.4\\
				\hline
			\end{tabular}
		\end{center}
		\label{tab:tsimp}
	\end{table}
	
	\subsection{Precipitation forecasting experiments}
	\begin{table}[]
		
		\caption{Performance on the REEC-2018 validation set. Note that ``p" denotes the initial probability to stop the back-propagation in STSGD.}
		\renewcommand{\arraystretch}{1.2}
		\begin{center}
			\footnotesize
			\begin{tabular}{c|c|c|c|c|c|c}
				\hline
				\multicolumn{2}{c|}{Model} & MSE & CSI & FAR & POD & COR\\
				\hline
				\multicolumn{2}{c|}{ConvLSTM~\cite{xingjian2015convolutional}} & 0.01156 & 0.5349 & 0.1733 & 0.5986 & 0.6851\\
				\hline
				\multirow{3}{*}{SCS}& p=0.0 & 0.01030 & 0.5624 & 0.1720 & \textbf{0.6372} & 0.7062\\
				& p=0.3 & 0.01033 & 0.5635 & 0.1702 & 0.6371 & \textbf{0.7072}\\
				& p=0.5 & \textbf{0.01022} & \textbf{0.5636} & \textbf{0.1682} & 0.6368 & \textbf{0.7072}\\
				\hline
			\end{tabular}
		\end{center}
		\label{tab:precipitationResult}
		\vspace{-0.2in}
	\end{table}
	
	The composite reflectance (CR) image received by the weather radar can reflect the precipitation situation in the specific area. By predicting the morphological changes of CR in the future we can forecast the precipitation. In this task, the models take a short period of the CR images as the input and generate the future CR images. The experiments are conducted on our REEC-2018 dataset which contains a set of CR images of Eastern China in 2018 and the CR image is recorded every 6 minutes. For better prediction, we select the top 100 rainy days from the dataset and crop a 224 $\times$ 244 pixel region as our input images. For preprocessing, we normalize the intensity value $Z$ of each pixel to $Z'$ by setting $Z'=\frac{Z-\min(\{Z_i\})}{\max(\{Z_i\}) - \min(\{Z_i\})}$, where $\{Z_i\}$ is the set of intensity values of all the pixels in the input image.
	
	In this task, we compare our SCS network with the ConvLSTM network. Both of them consist of an encoder and a decoder which have the same structure. For our model, the encoder and the decoder are 15-layer SCS networks while there are multi-stacked ConvLSTMs in the ConvLSTM version. Encoders take one frame in the CR sequence as input for every time-stamp and then generate the intermediate representation of the observed sequence. Decoders take the intermediate representation as well as the last CR image as input and generate the CR image prediction and new intermediate representation as shown in Fig.~\ref{fig:experiments} d.
	
	\begin{table*}[]
		\caption{Hyper-parameter settings for action recognition, outline annotation, auto driving and precipitation forecasting experiments.}
		\renewcommand{\arraystretch}{1.2}
		\begin{center}
			\footnotesize
			\begin{tabular}{c|c|c|c|c|c|c|c|c}
				\hline
				\multirow{2}{*}{} & \multicolumn{2}{c|}{Action Recognition} & \multicolumn{2}{c|}{Outline Annotation} & \multicolumn{2}{c|}{Auto Driving} & \multicolumn{2}{c}{Precipitation Forecasting}\\
				\cline{2-9}
				& BB supported & Stand-alone & LSTM & SCS & CNN+LSTM & SCS & ConvLSTM & SCS \\
				\hline
				Batch size & 16 & 40 & 8 & 4 & 128 & 128 & 8 & 4 \\
				Learning rate &1e-4 & 1e-4 & 1e-4 & 2e-4 & 2e-4 & 1e-4 & 1e-4 & 1e-4\\
				Backbone & \{VGG, Inception\} & - & VGG & VGG & ResNet-18~\cite{he2016deep}& - & - & -\\
				Num. layers & BB layers + 15 & 17 &\{2,5 10, 15\} & \{2,5 10, 15\} & 18 + 1& 15 & 15 & 15\\
				Training method &STSGD &STSGD &- &STSGD & -&ASTSGD & -& ASTSGD \\
				LTSC setting & $10\times 7$ & $5\times6$ & - & $10\times 4$ &- &- &- &- \\
				Feature dimension & 512 & 512 & 256 & 256 & 512 & 512 & 64 & 128 \\
				$\lambda$ & - & - & - & - &\{3, 6\} & \{3,6\}&- &- \\
				\hline
			\end{tabular}
		\end{center}
		\label{tab:hyperparameter}
		\vspace{-0.2in}
	\end{table*}
	
	This is a regression problem and every pixel of CR image represents the reflectance intensity of a specific geographic position. The networks are trained under the MSE loss function (the main goal $g$ is the MSE loss). The spatial goal $r_s$ is the same as the main goal. The temporal goal $r_t$ is to estimate the optical flow and pixel-wise difference between frames since every pixel has its own independent meaning: the reflectance intensity of that location. The optical flow guides $\mathcal{T}^2$ to learn the variation of wind direction while the pixel-wise difference is designed for the local precipitation changes. We evaluate the models using several metrics following~\cite{xingjian2015convolutional}, namely, mean squared error (MSE), critical success index (CSI), false alarm rate (FAR), probability of detection (POD) and correlation. Since every pixel has stand-alone meaning, we evaluate the performance at pixel level. We convert the prediction and the label to a 0/1 matrix using a threshold of 0.5 and define ``hit" (prediction=label=1), ``miss" (prediction=0, label=1), ``falsealarm" (prediction=1, label = 0). Then the metrics are defined as:
	\begin{align}
	& {\rm SCI}=\frac{\#{\rm hit}}{{\rm \#hit} + {\rm \#miss} + {\rm \#falsealarm}} \\
	& {\rm FAR}=\frac{{\rm \#falsealarm}}{{\rm \#hit}+{\rm \#falsealarm}}\\
	& {\rm POD}=\frac{{\rm \#hit}}{{\rm \#hit}+{\rm \#miss}}\\
	& {\rm correlation}=\frac{\sum_{i,j}{{\rm CR\_P}_{i,j}\times{\rm CR\_L}_{i,j}}}{\sqrt{(\sum_{i,j}{{\rm CR\_P}_{i,j}^2})(\sum_{i,j}{{\rm CR\_L}_{i,j}^2})+\epsilon}}
	\end{align}
	where ${\rm CR\_P}$ is the predicted CR image and ${\rm CR\_P}_{i,j}$ is the 0/1 value of position (i,j) in the CR image. ${\rm CR\_L}$ is the ground-truth CR image, i.e. the label.
	
	The models take 5 CR images as input and predict 5 future images. This is not a long sequence, so we do not adopt the LTSC scheme. STSGD is utilized in the deep SCS network. With a higher initial $p$, the model achieves better performance (see details in Tab.~\ref{tab:precipitationResult}), which indicates the important role of STSGD for SCS. The detailed comparison results are shown in Tab.~\ref{tab:precipitationResult}.

	\subsection{Optimization}
	The hyper-parameters are selected from grid searches and are listed in Tab.~\ref{tab:hyperparameter}. For all the experiments, the CNN layer is initialized with the ``Xavier initialization" method followed by Batch Normalization layer~\cite{ioffe2015batch}. All networks are trained using Adam optimizer~\cite{kingma2014adam} and the backbones are pre-trained on ImageNet. For the huge memory consumption of the long sequential vision tasks, the batch size of each training step is relative small and we accumulate the parameters' gradients of several training steps, then update the parameters together, which can speed up the training process to some extent. In the process of back-propagation-through-time (BPTT)~\cite{werbos1990backpropagation}, the gradients of RNN parameters was clipped to the range [-5, 5].

	\section{Data availability}
	The data that support the plots within this paper are available from the corresponding author upon reasonable request.
	
	\section{Code availability}
	A public version of the experiment codes will be made available with this paper, linked to from our website \href{http://www.mvig.sjtu.edu.cn}{http://www.mvig.sjtu.edu.cn} and \href{https://github.com/BoPang1996/Semi-Coupled-Structure-for-visual-sequental-tasks}{Github website}.

	\section{Author contributions}
	B.P. and C.L. conceived the idea. B.P., K.Z. and C.L. designed the experiments. B.P., K.Z., H.C., J.T. and M.Y. carried out programming, adjustment, and data analysis. B.P. and C.L. wrote the manuscript. B.P., J.T., M.Y. and all other authors contributed to the results analysis and commented on the manuscript.
	
	\section{Competing Interests} 
	The authors declare no competing interests.

\begin{thebibliography}{5}
		\bibitem{acuna2018efficient}
		Acuna, D., Ling, H., Kar, A \& Fidler, S.
		Efficient Interactive Annotation of Segmentation Datasets With Polygon-RNN++.
		In \textit{IEEE Conf. Comp. Vision and Pattern Recog.} 859--868 (2018).
		
		\bibitem{bottou2018optimization}
		Bottou, L., Curtis, F.E., \& Nocedal, J.
		Optimization methods for large-scale machine learning.
		\textit{SIAM Review} \textbf{60}, 223--311 (2018).
		
		\bibitem{carreira2017quo}
		Carreira, J. \& Zisserman, A.
		Quo vadis, action recognition? a new model and the kinetics dataset.
		In \textit{IEEE Conf. Comp. Vision and Pattern Recog.} 4724--4733 (2017).
		
		\bibitem{castrejon2017annotating}
		Castrejon, L., Kundu, K., Urtasun, R. \& Fidler, S.
		Annotating Object Instances with a Polygon-RNN.
		In \textit{IEEE Conf. Comp. Vision and Pattern Recog.} 2 (2017).
		
		\bibitem{chen2018lidar}
		Chen, Y. et al.
		Lidar-video driving dataset: Learning driving policies effectively.
		In \textit{IEEE Conf. Comp. Vision and Pattern Recog.} 5870--5878 (2018).
		
		\bibitem{cordts2016cityscapes}
		Cordts, M. et al.
		The cityscapes dataset for semantic urban scene understanding.
		In \textit{IEEE Conf. Comp. Vision and Pattern Recog.} 3213--3223 (2016).
		
		\bibitem{diez2015novel}
		Diez, I. et al.
		A novel brain partition highlights the modular skeleton shared by structure and function.
		\textit{Scientific reports} \textbf{5}, 10532 (2015).
		
		\bibitem{donahue2015long}
		Donahue, J. et al.
		Long-term recurrent convolutional networks for visual recognition and description.
		In \textit{IEEE Conf. Comp. Vision and Pattern Recog.} 2625--2634 (2015).
		
		
		\bibitem{feichtenhofer2018slowfast}
		Feichtenhofer, C., Fan, H., Malik, J. \& He, K.
		SlowFast networks for video recognition.
		In \textit{IEEE Int. Conf. Comp. Vision} 6202--6211 (2019).
		
		\bibitem{feichtenhofer2016convolutional}
		Feichtenhofer, C.,Pinz, A. \& Zisserman, A.
		Convolutional two-stream network fusion for video action recognition.
		In \textit{IEEE Conf. Comp. Vision and Pattern Recog.} 1933--1941 (2016).
		
		\bibitem{girdhar2019video}
		Girdhar, R., Carreira, J., Doersch, C. \& Zisserman, A.
		Video action transformer network.
		In \textit{IEEE Conf. Comp. Vision and Pattern Recog.} 244--253 (2019).
		
		\bibitem{graves2013generating}
		Graves, A.
		Generating sequences with recurrent neural networks.
		Preprint at https://arxiv.org/ abs/1308.0850 (2013).
		
		\bibitem{gu2017ava}
		Gu, C. et al.
		AVA: A video dataset of spatio-temporally localized atomic visual actions.
		In \textit{IEEE Conf. Comp. Vision and Pattern Recog.} 6047--6056 (2018).
		
		\bibitem{he2017mask}
		He, K., Gkioxari, G., Doll{\'a}r, P. \& Girshick, R.
		Mask r-cnn.
		In \textit{IEEE Int. Conf. Comp. Vision} 2980--2988 (2017).
		
		\bibitem{he2016deep}
		He, K., Zhang, X., Ren, S. \& Sun, J.
		Deep residual learning for image recognition.
		In \textit{IEEE Conf. Comp. Vision and Pattern Recog.} 770--778 (2016).
		
		\bibitem{hochreiter1997long}
		Hochreiter, S. \& Schmidhuber, J.
		Long short-term memory.
		\textit{Neural Computation} \textbf{9}, 1735--1780 (1997).
		
		\bibitem{hou2017tube}
		Hou, R., Chen, C. \& Shah, M.
		Tube convolutional neural network (T-CNN) for action detection in videos.
		In \textit{IEEE Int. Conf. Comp. Vision} 5822--5831 (2017).
		
		\bibitem{ioffe2015batch}
		Ioffe, S. \& Szegedy, C.
		Batch normalization: Accelerating deep network training by reducing internal covariate shift.
		In \textit{Int. Conf. Machine Learning} 448--456 (2015).
		
		\bibitem{ji20133d}
		Ji, S., Xu, W., Yang, M. \& Yu, K.
		3D convolutional neural networks for human action recognition.
		\textit{IEEE Trans. Pattern Analysis and Machine Intel.} \textbf{35}, 221-231 (2013).
		
		\bibitem{karpathy2014large}
		Karpathy, A. et al.
		Large-scale video classification with convolutional neural networks.
		In \textit{IEEE Conf. Comp. Vision and Pattern Recog.} 1725--1732 (2014).
		
		\bibitem{kim2017residual}
		Kim, J., El-Khamy, M. \& Lee, J.
		Residual LSTM: Design of a deep recurrent architecture for distant speech recognition.
		In \textit{Conf. Int. Speech Comm. Assoc.} 1591--1595 (2017).
		
		\bibitem{kingma2014adam}
		Kingma, D. \& Ba, J.
		Adam: A method for stochastic optimization.
		In \textit{Int. Conf. Learning Representations} (2015).
		
		\bibitem{kitamura2015entorhinal}
		Kitamura, T. et al.
		Entorhinal cortical ocean cells encode specific contexts and drive context-specific fear memory.
		\textit{Neuron} \textbf{87}, 1317--1331 (2015).
		
		\bibitem{krizhevsky2012imagenet}
		Krizhevsky, A., Sutskever, I. \& Hinton, G.
		Imagenet classification with deep convolutional neural networks.
		In \textit{Ann. Conf. Neural Inform. Proc. Sys.} 1097--1105 (2012).
		
		\bibitem{kuehne2011hmdb}
		Kuehne, H., Jhuang, H., Garrote, E., Poggio, T. \& Serre, T.
		HMDB: a large video database for human motion recognition.
		In \textit{IEEE Int. Conf. Comp. Vision} 2556--2563 (2011).
		
		\bibitem{levine2016end}
		Levine, S., Finn, C., Darrell, T. \& Abbeel, P.
		End-to-end training of deep visuomotor policies.
		\textit{J. Machine Learning Research} \textbf{17}, 1334--1373 (2016).
		
		\bibitem{lucas1986generalized}
		Lucas, B.D.
		\textit{Generalized image matching by the method of differences} (1986).
		
		\bibitem{maji2011action}
		Maji, S., Bourdev, L. \& Malik, J.
		Action recognition from a distributed representation of pose and appearance.
		In \textit{IEEE Conf. Comp. Vision and Pattern Recog.} 3177--3184 (2011).
		
		\bibitem{oliveri2009spatial}
		Oliveri, M., Koch, G. \& Caltagirone, C.
		Spatial--temporal interactions in the human brain.
		\textit{Experimental Brain Research} \textbf{195}, 489--497 (2009).
		
		\bibitem{pang2018deep}
		Pang, B., Zha, K., Cao, H., Shi, C. \& Lu, C.
		Deep RNN Framework for Visual Sequential Applications.
		In \textit{IEEE Conf. Comp. Vision and Pattern Recog.} 423--432 (2019).
		
		\bibitem{rumelhart1988learning}
		Rumelhart, D.E. et al.
		Learning representations by back-propagating errors.
		\textit{Cognitive modeling} \textbf{5} 1 (1988).
		
		\bibitem{santana2016learning}
		Santana, E. \& Hotz, G.
		Learning a driving simulator.
		Preprint at https://arxiv.org/abs/1608.01230 (2016).
		
		\bibitem{schulman2015trust}
		Schulman, J., Levine, S., Abbeel, P., Jordan, M. \& Moritz, P.
		Trust Region Policy Optimization.
		In \textit{Int. Conf. Machine Learning} 1889--1897 (2015).
		
		\bibitem{simonyan2014two}
		Simonyan, K. \& Zisserman, A.
		Two-stream convolutional networks for action recognition in videos.
		In \textit{Ann. Conf. Neural Inform. Proc. Sys.} 568--576 (2014).
		
		\bibitem{simonyan2014very}
		Simonyan, K., Zisserman, A.
		Very deep convolutional networks for large-scale image recognition.
		In \textit{Int. Conf. Learning Representations} (2015).
		
		\bibitem{song2017end}
		Song, S., Lan, C., Xing, J., Zeng, W. \& Liu, J.
		An end-to-end spatio-temporal attention model for human action recognition from skeleton data.
		In \textit{AAAI Conf. Art. Intel.} 4263--4270 (2017).
		
		\bibitem{soomro2012ucf101}
		Soomro, K., Zamir, A.R. \& Shah, M.
		UCF101: A dataset of 101 human actions classes from videos in the wild.
		Preprint at https://arxiv.org/abs/1212.0402 (2012).
		
		
		\bibitem{srivastava2015unsupervised}
		Srivastava, N., Mansimov, E. \& Salakhudinov, R.
		Unsupervised learning of video representations using lstms.
		In \textit{Int. Conf. machine learning} 843--852 (2015).
		
		\bibitem{sutskever2014sequence}
		Sutskever, I., Vinyals, O. \& Le, Q.V.
		Sequence to sequence learning with neural networks.
		In \textit{Ann. Conf. Neural Inform. Proc. Sys.} 3104--3112 (2014).
		
		\bibitem{szegedy2015going}
		Szegedy, C. et al.
		Going deeper with convolutions.
		In \textit{IEEE Conf. Comp. Vision and Pattern Recog.} 1--9 (2015).
		
		\bibitem{wang2011action}
		Wang, H., Kl{\"a}ser, A., Schmid, C. \& Liu C.
		Action recognition by dense trajectories.
		In \textit{IEEE Conf. Comp. Vision and Pattern Recog.} 443--455 (2011).
		
		\bibitem{wang2013dense}
		Wang, H., Kl{\"a}ser, A., Schmid, C. \& Liu, C.
		Dense trajectories and motion boundary descriptors for action recognition.
		\textit{Int. J. Comp. Vision} \textbf{103}, 60--79 (2013).
		
		\bibitem{wang2016actionness}
		Wang, L., Qiao, Y., Tang, X. \& Van G.L.
		Actionness estimation using hybrid fully convolutional networks.
		\textit{IEEE Conf. Comp. Vision and Pattern Recog.} 2708--2717 (2016).
		
		\bibitem{weinzaepfel2015learning}
		Weinzaepfel, P., Harchaoui, Z. \& Schmid, C.
		Learning to track for spatio-temporal action localization.
		In \textit{IEEE Int. Conf. Comp. Vision} 3164--3172 (2015).
		
		\bibitem{werbos1990backpropagation}
		Werbos, P.J. et al.
		Backpropagation through time: what it does and how to do it.
		\textit{Proceedings of the IEEE} \textbf{78}, 1550--1560 (1990).
		
		\bibitem{williams1990efficient}
		Williams, R.J. \& Peng, J.
		An efficient gradient-based algorithm for on-line training of recurrent network trajectories.
		\textit{Neural Computation} \textbf{2}, 490--501 (1990).
		
		\bibitem{wolman2012tale}
		Wolman, D.
		A tale of two halves.
		\textit{Nature} \textbf{483}, 260--263 (2012).
		
		\bibitem{wu2019long}
		Wu, C. et al.
		Long-term feature banks for detailed video understanding.
		In \textit{IEEE Conf. Comp. Vision and Pattern Recog.} 284--293 (2019).
		
		\bibitem{wu2015modeling}
		Wu, Z., Wang, X., Jiang, Y., Ye, H. \& Xue, X.
		Modeling spatial-temporal clues in a hybrid deep learning framework for video classification.
		In \textit{ACM Int. Conf. Multimedia} 461--470 (2015).
		
		\bibitem{xingjian2015convolutional}
		Shi, X. et al.
		Convolutional LSTM network: A machine learning approach for precipitation nowcasting.
		In \textit{Ann. Conf. Neural Inform. Proc. Sys.} 802--810 (2015).
		
		\bibitem{yue2015beyond}
		Yue-Hei N.J. et al.
		Beyond short snippets: Deep networks for video classification.
		In \textit{IEEE Conf. Comp. Vision and Pattern Recog.} 4694--4702 (2015).
		
		
		
	\end{thebibliography}
\end{document}